\documentclass[12pt]{article}
\usepackage[left=1in,right=1in,top=1in,bottom=1in,letterpaper]{geometry}

\usepackage{cite}
\usepackage{amsmath,amssymb,amsfonts}

\usepackage{algpseudocode}
\usepackage{algorithm}
\usepackage{graphicx}
\usepackage{textcomp}
\usepackage{xcolor}
 \usepackage{hyperref}
\newtheorem{theorem}{Theorem}[section]

\newtheorem{definition}[theorem]{Definition}

\newcommand{\norm}[1]{\left\lVert#1\right\rVert}
\newcommand{\R}{\mathbb{R}}

\DeclareMathOperator{\prox}{prox}

\DeclareMathOperator*{\argmin}{argmin}

\def\A{{\mathcal A}}

\def\J{{\mathcal J}}

\def\C{{\mathcal C}}
\def\B{{\mathcal B}}

\def\X{{\mathcal X}}
\def\Z{{\mathcal Z}}
\def\U{{\mathcal U}}

\def\S{{\mathcal S}}
\def\Y{{\mathcal Y}}

\graphicspath{{./pic}}
\usepackage{xcolor}

\def\BibTeX{{\rm B\kern-.05em{\sc i\kern-.025em b}\kern-.08em
    T\kern-.1667em\lower.7ex\hbox{E}\kern-.125emX}}

\providecommand{\keywords}[1]
{
  \small	
  \textbf{\textit{Keywords---}} #1
}

\begin{document}

\title{Hyperspectral Band Selection based on Generalized 3DTV and Tensor CUR Decomposition
\thanks{This research is partially supported by the NSF grant DMS-1941197.}
}

\author{Katherine Henneberger\thanks{Department of Mathematics, University of Kentucky, Lexington, KY 40511}
\and Jing Qin\thanks{Department of Mathematics, University of Kentucky, Lexington, KY 40511}}


\maketitle

\begin{abstract}
Hyperspectral Imaging (HSI) serves as an important technique in remote sensing. However, high dimensionality and data volume typically pose significant computational challenges. Band selection is essential for reducing spectral redundancy in hyperspectral imagery while retaining intrinsic critical information. In this work, we propose a novel hyperspectral band selection model by decomposing the data into a low-rank and smooth component and a sparse one. In particular, we develop a generalized 3D total variation (G3DTV) by applying the $\ell_1^p$-norm to derivatives to preserve spatial-spectral smoothness. By employing the alternating direction method of multipliers (ADMM), we derive an efficient algorithm, where the tensor low-rankness is implied by the tensor CUR decomposition. We demonstrate the effectiveness of the proposed approach through comparisons with various other state-of-the-art band selection techniques using two benchmark real-world datasets. In addition, we provide practical guidelines for parameter selection in both noise-free and noisy scenarios.
\end{abstract}


\section{Introduction}
Hyperspectral imaging (HSI) captures spectral information from ground objects across hundreds of narrow spectral bands. This method yields richer visual data than traditional RGB imagery and has found diverse applications in fields like remote sensing, biomedicine, agriculture, and art conservation. However, HSI data is characterized by its high dimensionality, leading to issues like spectral redundancy and increased computational demands. To address these challenges, it is crucial to apply dimensionality reduction and redundancy elimination strategies. The process of hyperspectral band selection is designed to mitigate these issues by choosing a subset of the original spectral bands, ensuring the retention of critical spectral information.

In general, HSI band selection can be implemented in an either supervised or unsupervised manner depending on the availability of label information. In this work, we focus on unsupervised band selection which does not require prior knowledge of labeled data. Many methods have been developed based on local or global similarity comparison. For example, the enhanced fast density-peak-based clustering (E-FDPC) algorithm \cite{jia2015novel} aims to identify density peaks to select the most discriminative bands, and the fast neighborhood grouping (FNGBS) method \cite{wang2020fast} utilizes spatial proximity and spectral similarity to group and select representative bands efficiently. Another notable approach is the similarity-based ranking strategy with structural similarity (SR-SSIM) \cite{xu2021similarity}, which prioritizes bands based on their contribution to the overall structural similarity, thus preserving the integrity of the spectral-spatial structures in HSI data.


Graph-based unsupervised band selection methods are gaining popularity for their ability to capture underlying graph similarities. Recently, the Marginalized Graph Self-Representation (MGSR) \cite{zhang2022marginalized} works to generate segmentation of the HSI through superpixel segmentation, thereby capturing the spatial information across different homogeneous regions and encoding this information into a structural graph. To accelerate the processing, the Matrix CUR decomposition-based approach (MHBSCUR) \cite{henneberger2023hyperspectral} has been developed. This graph-based method stands out for leveraging matrix CUR decomposition, which enhances computational efficiency. However, graph-based strategies entail extra computational overhead to construct graphs. Furthermore, most band selection approaches necessitate converting data into a matrix instead of directly applying algorithms on the HSI tensor. To address these issues, we propose a tensor-based HSI band selection algorithm that employs tensor CUR decomposition alongside a generalized 3D total variation (3DTV), which enhances the smoothness of the traditional 3DTV by applying the $\ell_p$-norm to derivatives.

Matrix CUR decompositions are efficient low-rank approximation methods that use a small number of actual columns and rows of the original matrix. Recent literature has extended the matrix CUR decomposition to the tensor setting. Initially the CUR decomposition was extended to tensors through the single-mode unfolding of 3-mode tensors, as discussed in early works \cite{mahoney2006tensor}. Subsequent research \cite{caiafa2010generalizing}, introduced a more comprehensive tensor CUR variant that encompasses all tensor modes. More recently, the terminology has been refined to include specific names for these decompositions, with \cite{cai2021mode,cai2021fast} introducing the terms Fiber and Chidori CUR decompositions. While these initial tensor CUR decomposition methods utilize the mode-$k$ product, a t-CUR decomposition has also been developed under the t-product framework \cite{chen2022tensor}.

In this paper, we focus on the t-product based CUR decomposition, which will be utilized to expedite a crucial phase in our proposed algorithm. This approach allows for the preservation of multi-dimensional structural integrity while facilitating a computationally efficient representation of the original tensor. This innovative combination of CUR and 3DTV in our algorithm leverages the inherent multidimensional structure of HSI data, making our method particularly suitable for dealing with large-scale datasets while ensuring robust band selection performance.

The organization of the remainder of this paper is as follows: Fundamental concepts and definitions in tensor algebra together with G3DTV are introduced in Section~\ref{sec:pre}. The proposed band selection method is presented in Section~\ref{sec:method}. Section~\ref{sec:exp} demonstrates the performance of the proposed method through several numerical experiments on benchmark remote sensing datasets. Conclusions and future work are discussed in Section~\ref{sec:con}.

\section{Preliminaries}\label{sec:pre}
In this section, we introduce fundamental definitions and notation about third-order tensors. Consider $\A \in \mathbb{R}^{n_1 \times n_2 \times n_3}$, $\B \in \mathbb{R}^{n_2 \times l \times n_3}$, and the $k$th frontal slice denoted by $\A_k = \A(:,:,k)$. We will use $\mathcal{O}$ to denote the zero tensor.
\begin{definition}
    The \textbf{block circulant operator} is defined as follows
\[
\text{bcirc}(\A) :=
\begin{bmatrix}
\A_1 & \A_{n_3} & \cdots & \A_2 \\
\A_2 & \A_1 & \cdots & \A_3 \\
\vdots & \vdots & \ddots & \vdots \\
\A_{n_3} & \A_{n_3-1} & \cdots & \A_1
\end{bmatrix}
\in \mathbb{R}^{n_1n_3 \times n_2n_3}.
\]
\end{definition}
\begin{definition} The operator \textbf{unfold($\cdot$)} and its inversion \textbf{fold($\cdot$)} for the conversion between tensors and matrices are defined as
\[
\text{unfold}(\A) =
\begin{bmatrix}
\A_1 \\
\A_2 \\
\vdots \\
\A_{n_3}
\end{bmatrix}
\in \mathbb{R}^{n_1n_3 \times n_2},
\quad
\text{fold}
\left(
\begin{bmatrix}
\A_1 \\
\A_2 \\
\vdots \\
\A_{n_3}
\end{bmatrix}
\right) = \A.
\]
\end{definition}

\begin{definition} The \textbf{block diagonal matrix form} of $\A$ is defined as a block diagonal matrix with diagonal blocks $\A^1,\ldots, \A^{n_3}$, i.e.,
$\text{bdiag}(\mathcal{A}) = \text{diag}(\A_1,\A_2,\ldots,\A_{n_3}).
$
\end{definition}

\begin{definition}
The  \textbf{t-product} of the tensors $\A$ and $\B$ is defined as
  $
  \A *\B := \text{fold} (\text{bcirc}(\A) \cdot \text{unfold}(\B)).
  $ The t-product can also be
converted to matrix-matrix multiplication in the Fourier
domain such that $\C = \A*\B$ is equivalent to bdiag$(\C_f) =\text{bcirc}(\A_f) \text{bcirc}(\B_f)$, where $\A_f = \verb|fft|(\A,[],3)$, the fast Fourier transform along the third dimension.
\end{definition}

\begin{definition}\label{def:orth}
A tensor $\mathcal{A}$ is \textbf{orthogonal} if $\mathcal{A}^**\mathcal{A}=\mathcal{A}*\mathcal{\A}^*=\J$, where $\J$ is the identity tensor such that $\J_f(:,:,i_3)=I_n$ for all ${i_3\in [n_3]}$.
\end{definition}

\begin{definition}\label{pseudoinv}
    The tensor $\A^\dag$ is the \textbf{Moore-Penrose pseudo inverse} calculated by finding the pseudo inverse of each face in the transform domain
    \[
    \A^\dag = \verb|ifft|\left(\verb|fold|\begin{bmatrix}
(\A_f)^\dag_1 \\
(\A_f)^\dag_2 \\
\vdots \\
(\A_f)^\dag_{n_3}
\end{bmatrix},
[],3\right).    \]
\end{definition}
\begin{definition} \label{def:fdiag}
A tensor $\A$ is \textbf{f-diagonal} if each frontal slice ${\A(:,:,i_3)}$ is diagonal for all  $i_3\in [n_3]$.
\end{definition}
\begin{definition}\label{def:tsvd}
The tensor \textbf{Singular Value Decomposition (t-SVD)} of a tensor induced by the t-product is
\[
\mathcal{\A}=\mathcal{U}*_L\mathcal{S}*_L\mathcal{V}^*\] where $\mathcal{U}\in\R^{n_1\times n_1\times n_3}$ and $\mathcal{V}\in\R^{n_2\times n_2\times n_3}$ are orthogonal and the core tensor $\mathcal{S}\in\R^{n_1\times n_2\times n_3}$ is f-diagonal.
Moreover, the \textbf{t-SVD rank} of tensor $\A$ is defined as
\[
\text{rank}_{t}(\A) = \#\{i: S(i,i,1)\neq 0\}\]
where $\#$ denotes the cardinality of a set.
\end{definition}

\begin{definition}\cite{chen2022tensor} Consider a tensor ${\Y\in \R^{n_1\times n_2\times n_3}}$. Let ${I\subset\{1:n_1\}}$ and ${J\subset\{1:n_2\}}$ be index subsets and define ${\mathcal{R} = \Y(I,:,:)}$, ${\C = \Y(:,J,:)}$, and ${\U = \Y(I,J,:)}$. The \textbf{t-CUR decomposition} of $\Y$ is $\hat{\Y}=\C* {\U}^\dagger*\mathcal{R}$, where $\U^\dagger$ is the pseudoinverse of $\U$ as defined in Definition \ref{pseudoinv} above.

\end{definition}
\begin{definition}
The \textbf{proximal operator} of  $\ell_1^p$ is defined as
\begin{align}\label{eqn:prox}
    \prox_{\lambda \norm{\cdot}_1^p}(\Z) = \argmin_{\X\in\R^{n_2\times l\times\ldots\times n_m}}\frac{1}{2}\|\X-\Z\|^2+ \lambda \|\X\|_1^p.
\end{align}
\end{definition}
Note that the proximal operator $\prox_{\lambda \norm{\cdot}_1^p}(\Z)$ when $p=1,2,3,4$ can be computed using Algorithms 1 and 2 in \cite{prater2023constructive}.

In addition, to describe the high order smoothness of 3D images, we propose a novel 3D total variation regularization.
\begin{definition}\label{def:g3dtv}
    Consider a function ${u:\Omega\subset\mathbb{R}^3\to\mathbb{R}}$ where $\Omega$ is compact, open and bounded. The generalized 3D total variation (G3DTV) regularization of $u$ is defined as
    \[
\norm{u}_{G3DTV}=\sum_{i=1}^3\norm{\nabla_i u}_1^p,
    \]
    where $\nabla_iu$ is the derivative along the $i$th axis and $p\geq1$ is an integer. If $\mathcal{U}$ is a discretized version of $u$ when $\Omega$ is discretized as a grid, then $\mathcal{U}$ becomes a third-order tensor and thereby the definition can be extended to a discrete case.
\end{definition}

\section{Proposed method}\label{sec:method}
Consider a hyperspectral data tensor $\Y\in\mathbb{R}^{n_1\times n_2\times n_3}$ with $n_3$ spectral bands, each composed of $n_1\times n_2$ spatial pixels. To select bands, we aim to decompose $\Y$ into a sum of low-rank and spatial-spectral smooth tensor $\B$ and a sparse tensor $\S$ via the following model
\begin{equation}\label{eqn:model}
\min_{\text{rank}_{t}(\B)\leq r,\S}\frac12\norm{\B+\S-\Y}_F^2+\lambda_1\norm{\S}_1+\lambda_2\norm{\mathcal{B}}_{G3DTV}.
\end{equation}
Here the third term is the G3DTV of $\mathcal{B}$ as defined in Definition~\ref{def:g3dtv}, i.e., $\norm{\mathcal{B}}_{G3DTV}=\sum_{i=1}^3\norm{\nabla_i\B}^p_1$.
The parameter $\lambda_1$ is the regularization parameter for controlling the sparsity of the outlier tensor $S$, and  $\lambda_2>0$ controls the spatial-spectral smoothness of $\B$. The parameter $p$ in G3DTV is an integer, which is set as 2 throughout our experiments. In order to apply the ADMM framework to minimize \eqref{eqn:model}, we introduce the auxiliary variables $\X_i$ and rewrite \eqref{eqn:model} as an equivalent form
\begin{equation}
\begin{aligned}
\min_{\substack{\text{rank}_t(\B)\leq r\\\S,\X}}&\frac12\norm{\B+\S-\Y}_F^2+\lambda_1\norm{\S}_1+\lambda_2\sum_{i=1}^3\norm{\X_i}^p_1\\
&\text{s.t.}\quad\X_i=\nabla_i\B\quad \text{for}\quad i=1,2,3.
\end{aligned}
\end{equation}
We introduce an indicator function to take care of the constraints. Let $\Pi=\{\X\in\R^{n_1\times n_2\times n_3}\,|\, \text{rank}_{\text{t-SVD}}(\X)\leq r\}$. The indicator function $\chi_\Pi$ is defined as
 $\chi_{\Pi}(\X)=0$ if $\X\in\Pi$ and $\infty$ otherwise. Then the augmented Lagrangian reads as
\[\begin{aligned}\label{eqn:lagrange}
\mathcal{L}=&\frac12\norm{\B+\S-\Y}_F^2+\lambda_1\norm{\S}_1 + \chi_\Pi(\B)+\lambda_2\sum_{i=1}^3\norm{\X_i}^p_1\\
&+\frac{\beta}2\sum_{i=1}^3\norm{\nabla_i\B-\X_i+\widetilde{\X}_i}_F^2,
\end{aligned}\]
where $\widetilde{\X}$ is a dual variable and $\beta>0$ is the penalty parameter.
The resulting algorithm can be described as follows:
\[
\left\{
\begin{aligned}
    \B\leftarrow& \argmin_{\text{rank}_{t}(\B)\leq r}\frac12\norm{\B+\S-\Y}_F^2\\&+\frac{\beta}2\sum_{i=1}^3\norm{\nabla_i\B-\X_i+\widetilde{\X}_i}_F^2,\\
    \S\leftarrow&\argmin_{\S}\frac12\norm{\B+\S-\Y}_F^2+\lambda_1\norm{\S}_1,\\
    \X_i\leftarrow&\argmin_{\X_i}\lambda_2\norm{\X_i}^p_1+\frac{\beta}2\norm{\nabla_i\B-\X_i+\widetilde{\X}_i}_F^2,\\
    \widetilde{\X}_i\leftarrow& \widetilde{\X}_i+\nabla_i\B-\X_i,
    \end{aligned}\right.
\]
where $\X_i$ and $\widetilde{\X}_i$ are updated for $i=1,2,3$. Applying the ADMM algorithm requires solving three subproblems at each iteration. Specifically, we aim to minimize $\mathcal{L}$ with respect to $\B$, $\S$, and $\X_i$. A common approach for updating $\B$ is to use the skinny t-SVD, however this can be costly when the size of the tensor is large. To address this issue, we utilize the t-CUR decomposition and update $\B$ via
\[
\begin{aligned}
   \B^{j+1}&= \argmin_{\text{rank}_{t}(\B)\leq r}\frac12\norm{\B+\S^j-\Y}_F^2\\
   &+\frac{\beta}2\sum_{i=1}^3\norm{\nabla_i\B-\X^j_i+\widetilde{\X}^j_i}_F^2:=f(\B).
\end{aligned}
\]
By applying gradient descent with step size $\tau>0$, we obtain the updating scheme for $\mathcal{B}$ as
$\B^{j+1}=\B^j-\tau\nabla f(\B^j)$ where the gradient is calculated as
$\nabla f(\B^j)=\B^j+\S^j-\Y+\beta\sum_{i=1}^3\nabla_i^T(\nabla_i\B^j-\X^j_i+\widetilde{\X}^j_i)$.
Then the t-CUR decomposition is updated as
\[
\left\{\begin{aligned}
\C\leftarrow& \C-\tau\nabla f(\B^j)(:,J,:),\\
\mathcal{R}\leftarrow& \mathcal{R}-\tau\nabla f(\B^j)(I,:,:),\\
\U\leftarrow& \tfrac12(\C+\mathcal{R}),
\end{aligned}
\right.\]
where $I$ and $J$ are the respective row and column index sets. Next we calculate $\U^\dag$ using Definition~\ref{pseudoinv}
\begin{align}\label{Bupdate}
    \B^{j+1} = \C*\U^{\dag}*\mathcal{R}.
\end{align}
By fixing other variables, we update $\S$ and $\X_i$ as
\begin{align}
S^{j+1}
&=\prox_{\frac{\lambda_1}{\beta}\norm{\cdot}_1}(\Y-\B^{j+1})\label{Supdate}\\
\X_i^{j+1}&=\prox_{\frac{\lambda_2}{\beta}\norm{\cdot}^p_1}(\nabla_i\B^{j+1}+\widetilde{\X}^j_i),\quad i=1,2,3.\label{Xupdate}
\end{align}
Here $\prox_{\frac{\lambda_1}{\beta}\norm{\cdot}_1}$ is the soft thresholding operator and $\prox_{\frac{\lambda_2}{\beta}\norm{\cdot}^p_1}$ is the $\ell_1^p$ proximal operator defined in \eqref{eqn:prox}.
The convergence condition is defined as:
${\norm{\B^{j+1}-\B^{j}}_\infty<\varepsilon}$, where $\varepsilon$ is a predefined tolerance. Finally, we apply a classifier such as k-means on $\B^{j+1}$ to find the desired $k$ clusters. The fiber indices of the bands closest to the cluster centroids are stored in the set $Q$. Thus the corresponding bands from the original tensor $\Y$ represent the desired band subset.

The main computational cost of Algorithm \ref{alg:cap} arises from updating $\B$ using the t-product. The per iteration complexity is $O(2n_3\log(n_3)+n_1n_2\max(s_r,s_3)n_3)$. Thus a computational trade-off exists when applying the t-CUR decomposition.

\begin{algorithm}
\caption{Hyperspectral Band Selection Based on Tensor CUR Decomposition}\label{alg:cap}
\begin{algorithmic}
\State\textbf{Input:} $\Y\in\R^{n_1\times n_2\times n_3}$, maximum number of iterations $T$, number of sampled rows and columns $s_r$ and $s_c$, number of desired bands $k$, parameters $\lambda_1,\lambda_2, \beta,$ and tolerance $\varepsilon$
\State\textbf{Output:} The index $Q$ of the desired band set.
\State 1. Optimize the model in (\ref{eqn:model}) using ADMM:
    \State\textbf{Initialize:} $\B^0,\S^0,\X_i^0=\mathcal{O}$
    \For{$j=0,1,2,3, \ldots T$ }
    \State Update $\B^{j+1}$ as in (\ref{Bupdate})
    \State Update $\S^{j+1}$ by solving (\ref{Supdate})
    \State Update $\X_i^{j+1}$ by solving (\ref{Xupdate})
    \State Update $\widetilde{\X}_i^{j+1}=\widetilde{\X}^j_i+\nabla_i\B^{j+1}-\X^{j+1}_i,\quad i=1,2,3$
    \State Check the convergence conditions
    \If{converged}
    \State Exit and set $\widetilde{\B}=\B^{j+1}$
    \EndIf
    \EndFor
    \State Set $\widetilde{\B}= \B^{T+1}$
\State 2. Cluster faces of $\widetilde{\B}$ using k-means and find the index set $Q$ which indicates the bands closest to the $k$ cluster centroids.
\end{algorithmic}
\end{algorithm}

\section{Experimental data and results}\label{sec:exp}

\subsection{Experimental Setup}

In our numerical experiments, we evaluate the proposed method using two publicly available HSI datasets: Indian Pines and Salinas-A. To assess the effectiveness of our approach, we compare it with several other state-of-the-art band selection methods, including E-FDPC \cite{jia2015novel}, SR-SSIM \cite{xu2021similarity}, and FNGBS \cite{wang2020fast} MGSR \cite{zhang2022marginalized}, and MHBSCUR \cite{henneberger2023hyperspectral}. To assess these method's performance, we conduct classification tests using support vector machine (SVM), $k$-nearest neighborhood (KNN), and convolutional neural network (CNN) as classifiers. Then the overall accuracy (OA) is used as the classification metric. According to our experiments, the classifiers such as SVM and KNN yield better classification accuracy and are faster across all the band selection methods than CNN. Therefore, we only report the SVM and KNN overall accuracy results in this paper. For the classification setup, we randomly select $90\%$ of the samples from each dataset for training and  the remaining $10\%$ for testing. Each classification test is repeated 50 times with different training data to reduce the randomness effect. We test the number of bands ranging from 3 to 30, increasing in increments of three.

For Algorithm~\ref{alg:cap}, the selection of specific rows and columns is fixed, with the actual choices being made through a random permutation initialized by setting the Mersenne Twister generator's seed to 1. For each dataset, classification approach, and targeted number of bands, we fine-tune the parameters $\lambda_1,\lambda_2$, $\beta$, and $\tau$ using a grid search. The parameters $\lambda_1$ and $\lambda_2$ are adjusted over the range $\{10^{-4}, 10^{-3}, 10^{-2}, 10^{-1}, 1, 10, 100, 1000\}$. The parameter $\beta$ is tuned over the range $\{10^{-1},  1, 10, 100\}$ and $\tau$ is similarly optimized over the set $\{10^{-1}, 1, 10, 100\}$.

All the numerical experiments are conducted using MATLAB 2023b on a desktop computer equipped with an Intel i7-1065G7 CPU, 12GB RAM, and running Windows 11.

\subsection{Experiment 1: Indian Pines}
\begin{figure}[ht]
    \centering      \includegraphics[width=.35\textwidth]{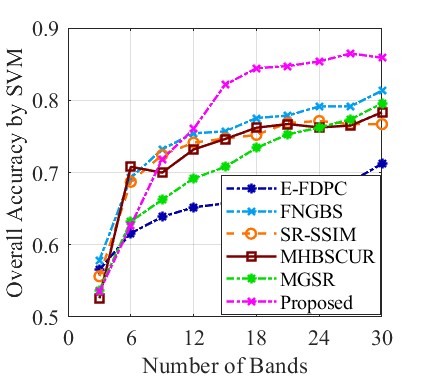}
    \vspace{-10pt}
\caption{Overall Accuracy with SVM for Indian Pines}
\label{fig:IndianPinesOASVM}
\end{figure}

\begin{figure}[ht]
    \centering      \includegraphics[width=.35\textwidth]{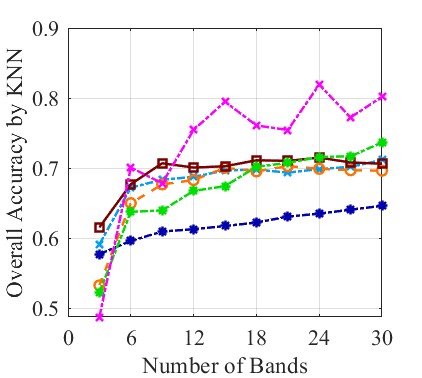}
        \vspace{-10pt}
\caption{Overall Accuracy with KNN for Indian Pines}
\label{fig:IndianPinesOAKNN}
\end{figure}

In the first experiment, we test the Indian Pines dataset\footnote{
\href{{https://www.ehu.eus/ccwintco/index.php/Hyperspectral\_Remote\_Sensing \_Scenes\#Indian\_Pines}}{https://www.ehu.eus/ccwintco/index.php/Hyperspectral\_Remote\_Sensing \_Scenes\#Indian\_Pines}}. The dataset, acquired by the AVIRIS sensor in Northwestern Indiana, contains 145x145 pixels, 200 spectral bands, and 16 distinct classes. Fig. \ref{fig:IndianPinesOASVM} and Fig. \ref{fig:IndianPinesOAKNN} plot the OA curves produced by SVM and KNN respectively for this dataset. The proposed method (THBSCUR) outperforms the state-of-the-art methods in terms of OA when SVM or KNN is used to classify the results for a high number of bands. In this case, SVM also produces slightly higher classification accuracy than KNN. The average running times for each method in seconds averaged over the number of selected bands $k$ are $0.0382$ for E-FDPC, $0.1417$ for FNGBS, $30.1334$ for SR-SSIM, $15.3760$ for MHBSCUR, $19.7766$ for MGSR, and $169.5353$ for our proposed method. One can see that While our method may require a longer running time, it consistently achieves superior performance, particularly with a larger number of bands.

\subsection{Experiment 2: Salinas-A}
\begin{figure}
    \centering    \includegraphics[width=.35\textwidth]{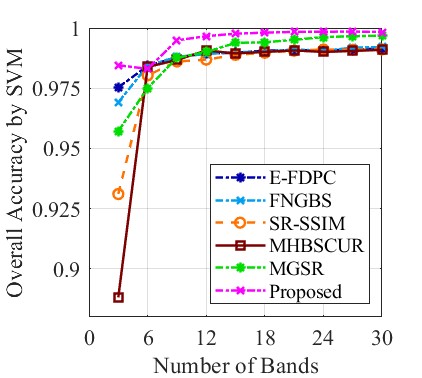}
        \vspace{-8pt}
\caption{Overall Accuracy by SVM for Salinas-A }
\label{fig:SalinasOASVM}
\end{figure}

\begin{figure}
    \centering    \includegraphics[width=.35\textwidth]{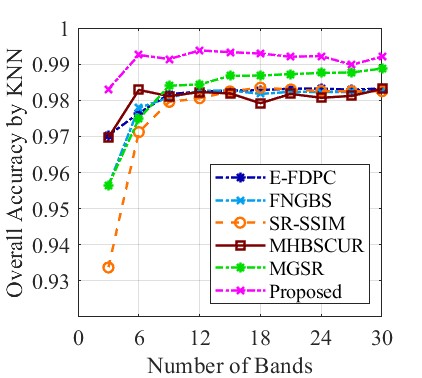}
        \vspace{-8pt}
\caption{Overall Accuracy by KNN for Salinas-A}
\label{fig:SalinasOAKNN}
\end{figure}

In the second experiment, we test the Salinas-A dataset, which is a subset of a larger image captured by the AVIRIS sensor in California, consisting of $86\times 83$ pixels, $204$ bands and $6$ classes. Fig. \ref{fig:SalinasOASVM} and Fig. \ref{fig:SalinasOAKNN} plot the OA curves produced by SVM and KNN respectively for the Salinas-A dataset. The average running times for each method in seconds averaged over the number of selected bands $k$ are $0.0107$ for E-FDPC, $0.0502$ for FNGBS, $22.673$ for SR-SSIM, $6.1636$ for MHBSCUR, $5.9491$ for MGSR, and $75.1455$ for our proposed method. Despite longer running time, the proposed method exhibits consistently superior performance with both the SVM and KNN classifier.

\section{Discussion}
In this section, we discuss parameter selection for each test dataset and provide general guidelines.  Table \ref{tab:parameters} lists the optimal parameter range for each parameter and dataset. As the number of selected bands  increases, the optimal choices for $\lambda_1$ and $\lambda_2$ are stable, whereas parameters $\beta$ and $\tau$ must be more carefully tuned. Importantly, there exist multiple combinations of parameter choices that may lead to optimal overall accuracy.

\begin{table}
    \caption{Parameter range for each dataset}
    \label{tab:parameters}
    \centering
    \begin{tabular}{c|c|c}
    \hline\hline
        Parameters & Indian Pines& Salinas-A \\
         \hline
$\lambda_1 $& $\{10^{-4},10^{-3}\}$ &$\{10^{-3}\}$ \\
        $\lambda_2$&$\{10^{-4},10^{-3},10^{-2}\}$ &$\{10^{-3}\}$\\
        $\beta$ &$\{0.1,1,10\}$&$\{0.1,1,10,100\}$\\
        $\tau$ &$\{1,10,100\}$  &$\{0.1,1\}$\\ \hline\hline
    \end{tabular}
\end{table}

In addition to the optimal parameter ranges outlined in Table \ref{tab:parameters}, it is important to consider the effects of noise on parameter selection. Hyperspectral data often contains noise, commonly modeled as Gaussian noise, can greatly impact the performance of band selection algorithms. To account for the noise, the parameters may need to be adjusted to accommodate the deviation from the original signal introduced by the noise.

Considering a scenario where Gaussian noise with various standard deviation levels is introduced into the dataset we recommend the following guidelines for selecting 15 bands from the Indian Pines dataset. The parameters $\lambda_1$ and $\lambda_2$, which control the trade-off between the sparsity term and the 3DTV regularization term, may require adjustment to counteract the introduction of noise. For the noisy case, $\beta=1$ and $\tau=10^{-4}$ consistently yield optimal results across noise levels. Typically, a higher level of noise may necessitate a larger value of $\lambda_2$ as it enforces stricter smoothness in $\B$. The parameter $\lambda_1$ may be chosen slightly higher than typical settings in noise-free environments. The optimal parameters for $\sigma = 1,2,3$ are presented in \ref{tab:noisyparameters}.

\begin{table}
    \caption{Optimal parameters for Indian Pines with Gaussian noise}
    \label{tab:noisyparameters}
    \centering
    \begin{tabular}{c|c|c|c}
    \hline\hline
        Parameters & $\sigma = 1$ & $\sigma = 3$ & $\sigma = 5$  \\
         \hline
        $\lambda_1 $    & $10^{-2}$ &$10^{-3}$  & $10^{-2}$\\
        $\lambda_2$     & $0.1$     &$1$        & $0.1$ \\
        $\beta$         & $1$       &$1$        & $1$  \\
        $\tau$          & $10^{-4}$ &$10^{-4}$  & $10^{-4}$ \\ \hline\hline
    \end{tabular}
\end{table}

\section{Conclusion}\label{sec:con}
The inherent high dimensionality and redundancy of HSI data necessitates effective strategies for band selection to ensure computational efficiency and data interpretability. In this work, we propose a novel tensor-based band selection approach that utilizes the G3DTV regularization to preserve high-order spatial-spectral smoothness, and the tensor CUR decomposition to improve processing efficiency while preserving low-rankness. Our method distinguishes itself by maintaining the tensor structure of HSI data, avoiding the often inefficient conversion to matrix form and directly addressing the spectral redundancy problem in the tensor domain. This approach not only preserves the multidimensional structure of the data but also enhances computational scalability, making it ideal for handling large-scale datasets in practical applications. In future work, we aim to further improve the efficiency of our algorithm by developing an accelerated version. Furthermore, we plan to explore various importance sampling schemes in the t-CUR decomposition and study the impact of gradient tensor sparsity on the band selection accuracy.

\bibliographystyle{unsrt}

\bibliography{references}

\end{document}